\documentclass[manuscript,screen]{acmart}

\AtBeginDocument{%
  \providecommand\BibTeX{{%
    \normalfont B\kern-0.5em{\scshape i\kern-0.25em b}\kern-0.8em\TeX}}}

\definecolor{lightblue}{RGB}{212, 235, 255}
\definecolor{orange}{RGB}{255, 105, 0}
\definecolor{lightgreen}{RGB}{177, 231, 171}
\definecolor{lightyellow}{RGB}{255, 255, 148}

\usepackage{array}
\newcolumntype{P}[1]{>{\centering\arraybackslash}p{#1}}
\newcolumntype{L}[1]{>{\raggedright\let\newline\\\arraybackslash\hspace{0pt}}p{#1}}
\newcolumntype{R}[1]{>{\raggedleft\arraybackslash}p{#1}}

\usepackage{float}
\usepackage{enumitem}

\newcommand{\rednote}[1]{{\color{red}#1}}

\usepackage{tcolorbox}

\setcopyright{none}
\renewcommand\footnotetextcopyrightpermission[1]{} 
\begin{document}




\title{NLP for Maternal Healthcare: \newline Perspectives and Guiding Principles in the Age of LLMs}

\author{Maria Antoniak}
\affiliation{%
  \institution{Allen Institute for AI}
  \country{USA}
}

\author{Aakanksha Naik}
\affiliation{%
  \institution{Allen Institute for AI}
  \country{USA}
}

\author{Carla S. Alvarado}
\affiliation{%
  \institution{Association of American Medical Colleges, Center for Health Justice}
  \country{USA}
}

\author{Lucy Lu Wang}
\affiliation{%
  \institution{University of Washington, Allen Institute for AI}
  \country{USA}
}

\author{Irene Chen}
\affiliation{%
  \institution{University of California, Berkeley and University of California, San Francisco}
  \country{USA}
}

\renewcommand{\shortauthors}{Antoniak et al.}

\begin{abstract}
\rednote{This is an unpublished pre-print and has not been peer-reviewed. Before citing, please check authors’ websites for updated version and publication information. Comments welcome.}

\vspace{4mm}

Ethical frameworks for the use of natural language processing (NLP) are urgently needed to shape how large language models (LLMs) and similar tools are used for healthcare applications. 
Healthcare faces existing challenges including the balance of power in clinician-patient relationships, systemic health disparities, historical injustices, and economic constraints. 
Drawing directly from the voices of those most affected, and focusing on a case study of a specific healthcare setting, we propose a set of guiding principles for the use of NLP in maternal healthcare.
We led an interactive session centered on an LLM-based chatbot demonstration during a full-day workshop with 39 participants, and additionally surveyed 30 healthcare workers and 30 birthing people about their values, needs, and perceptions of NLP tools in the context of maternal health. 
We conducted quantitative and qualitative analyses of the survey results and interactive discussions to consolidate our findings into a set of guiding principles.
We propose nine principles for ethical use of NLP for maternal healthcare, grouped into three themes: (i) recognizing contextual significance
(ii) holistic measurements, and (iii) who/what is valued. 
For each principle, we describe its underlying rationale and provide practical advice. 
This set of principles can provide a methodological pattern for other researchers and serve as a resource to practitioners working on maternal health and other healthcare fields to emphasize the importance of technical nuance, historical context, and inclusive design when developing NLP technologies for clinical use.
\end{abstract}




\maketitle

As natural language processing (NLP) methods and large language models (LLMs) have increased in size and performance, so has hype and excitement increased about their clinical use~\citep{thirunavukarasu2023large}. 
Recent work experimenting with generative LLMs has found promising results in comparison to physicians, including for tasks such as addressing public health concerns \citep{ayers2023evaluating}, answering care-seekers' questions \citep{ayers2023comparing}, and engaging in diagnostic dialogues \citep{tu2024conversational}.
But healthcare and ethics researchers have also highlighted the safety risks \citep{mehandru2022reliable}, biases \citep{xiao2023name}, inaccuracies \citep{boag2021pilot}, and other problems arising from NLP tools applied to healthcare contexts, as well as general risks of LLMs \citep{bender2021dangers,weidinger2022taxonomy}.
Given the sensitivity of medical data and the potential for harmful negative outcomes, deciding when and how to employ NLP technologies in clinical settings requires careful consideration.

Two key challenges complicate these decisions.
First, numerous stakeholders have a shared interest in the development and use of these technologies; care-seekers, clinicians, researchers, hospital administration, and other groups can all provide design and decision-making input, but these voices are not equally represented, may conflict, and are rarely brought together during decisions around system design and implementation. 
Second, most research developing ethical guidelines has studied high level applications of machine learning for healthcare, across (i) a range of medical topics and (ii) across a range of machine learning methods.
There is a gap for research grounded in a specific healthcare setting where risks, benefits, and stakeholder perceptions can be fully explored, focusing specifically on NLP technologies and the new risks posed by LLMs.

In this work, we focus on a case study of \textit{maternal healthcare} in the United States. 
We employ a participatory design framework~\citep{muller1993participatory} to amplify the voices of diverse stakeholder groups who have previously been underrepresented in the discussion around use of NLP technologies in healthcare.
Drawing on the input from these groups, we identify a set of \textit{guiding principles} that support well-grounded and ethical uses of NLP and LLM technologies in maternal healthcare.

We choose maternal health for three reasons. 
First, there are many prior research studies and applications of NLP methods focused on maternal healthcare \citep{lee-etal-2021-classification,goodrum-etal-2019-extraction,knowles2014high,valdeolivar2023towards}.
Second, pregnancy and childbirth are common events that often comprise a person's sole or major interaction with the healthcare system, increasing the significance and also abundance of perspectives on this topic.
Third, maternal health is a ``perfect storm'' of healthcare vulnerabilities, with historical biases and power dynamics influencing care; for example, maternal health in the U.S.~has received much attention in recent years due to the high morbidity and mortality of birthing people and significant racial inequities \citep{Creanga2015-on,Martin2017-nothing}.
Focusing on maternal health allows us to map our investigation of NLP applications over a set of \emph{specific} and \textit{grounded} opportunities and risks.

We solicited key considerations about an LLM-based chatbot and NLP tools from diverse stakeholders, including clinicians, birthing people, researchers, community health workers, government and non-profit workers, and others.
We collected perspectives in two main ways (Figure~\ref{fig:overview}).
First, we introduced stakeholders to NLP technologies, allowing them to surface concerns, opportunities, and discussion through interaction with these technologies in a controlled setting.
Second, we conducted post-interaction surveys to gather information about priorities, opportunities, and risks.

Through analysis of the collected data, iterative rounds of discussion, and literature review, we consolidated and identified nine guiding principles for the use of NLP in maternal healthcare, organized according to three themes: \textit{context}, \textit{measurements}, and \textit{values}. 
These themes are inspired by our maternal health case study but provide a transferable blueprint for mapping tensions in other healthcare areas, supporting NLP practitioners and healthcare administrators on decisions related to the design and deployment of NLP tools. 

\section{Related Work}
\label{section:background}

\subsection{NLP and LLMs for Healthcare} 

NLP methods have been applied to many kinds of health data, both in research and in industry applications.
Traditionally, NLP for healthcare has focused on structured tasks like information extraction from clinical notes \citep{Savova2010MayoCT, Wang2018ClinicalIE, Si2019EnhancingCC} and outcome prediction \citep{Shamout2020MachineLF, naik-etal-2022-literature, van-aken-etal-2021-clinical}, or on public health surveillance of social media \citep{Nikfarjam2015PharmacovigilanceFS, LeGlaz2019MachineLA, Baclic2020ChallengesAO}.
Technical advances in LLMs have shifted this focus, with recent work applying generative NLP tools to a wide variety of question-answering use cases, including engaging in diagnostic dialogues \citep{tu2024conversational}, addressing medical scenarios~\cite{nori2023capabilities,ueda2023evaluating}, and generating responses for patients asking questions on social media sites~\cite{ayers2023comparing}.
Research has also focused on developing large datasets~\citep{fleming2023medalign} or studying clinician adoption of LLMs~\cite{sivaraman2023ignore,kim2023organizational}.
On the applied side, as one example, Epic Systems, the leading provider of electronic health record (EHR) systems in the United States, is bringing GPT-4 into EHRs to help clinicians communicate with patients~\citep{epicehr}. 
And in practice, everyday users are already exploring their healthcare questions with generic LLM-based chat tools; according to one analysis of a publicly-released dataset of user-GPT conversation histories, approximately 3\% of user queries are health-related \citep{ouyang2023shifted}.

These use cases have fueled both optimism~\citep{vaishya2023chatgpt,lee2023ai} and  scrutiny~\citep{wornow2023shaky,vaishya2023chatgpt}. 
Within the fairness, accountability, and transparency (FAccT) community, researchers have studied many general risks and challenges arising with LLMs \citep{bender2021dangers,weidinger2022taxonomy} and also highlighted risks specific to healthcare settings.
For example, recent work on ethical uses of machine learning for healthcare has studied bias measurement \citep{xiao2023name}, explainability methods \citep{panigutti2020doctor,balagopalan2022road,poulain2023improving}, dataset documentation gaps \citep{rostamzadeh2022healthsheet}, and procedures for ethical implementations and deployments \citep{sendak2020human}.
Compared to research on other application areas, these works emphasize the particular privacy issues associated with clinical data \citep{xiao2023name}, the importance of optimizing for the right outcome variables (health outcomes rather than institutional costs) \citep{obermeyer2019dissecting}, and interactions between healthcare workers and AI systems \citep{fogliato2022who} that require explainability \citep{panigutti2020doctor,balagopalan2022road,poulain2023improving}.

Most of these works addressing healthcare contexts study machine learning methods more broadly, rather than focusing on NLP methods and the specific challenges arising from new tools like LLMs.
We build on these works by focusing on NLP tools for maternal health and by eliciting perceptions from multiple stakeholder groups.

\subsection{Ethical Guidelines for NLP/AI and Healthcare} 

Prior work has developed important guidelines and frameworks for the use of machine learning (ML) and artificial intelligence methods for healthcare.
These works take a general view of healthcare
and are often focused on ML practitioners and their processes.
For example, \citet{mccradden2023whats} focuses on ethical guidelines for ML-informed clinical decision-making, and \citet{chen2021ethical} and \citet{Wiens2019DoNH} provide overviews of the ML/NLP development pipeline and recommendations focused on each pipeline step.
The recommendations in these works are based on literature reviews and broad sets of healthcare examples.
While most of these guidelines consider healthcare broadly, \citet{sendak2020human} design guidelines based on the deployment of a specific sepsis-detection machine learning tool, and \citet{petti2023ethical} focused on developing ethical guidelines for the use of NLP and AI methods for early detection of Alzheimer's disease.
Our work builds most directly on the latter study, as we also focus on linguistic data and NLP methods.

We build on these works by combining in a novel way the following goals: (i) focusing on NLP and new LLM technologies, (ii) grounding our work in maternal health as a specific healthcare context, and (iii) soliciting direct feedback from multiple stakeholder groups.
In particular, we elicit values from various stakeholders using a framework introduced by \citet{jakesch2022different}, which aims to model the ``human process'' at the center of healthcare \citep{Ghassemi2022MachineLA}.
As recommended by \citet{rajkomar2018ensuring} and following prior work \cite{vedam2019giving}, we follow a research process that draws on \textit{participatory design}, a collaborative design approach that involves stakeholders and end users in the design process for new technologies~\cite{muller1993participatory}, such that their needs drive design decisions and result in tools that better address their concerns and challenges~\cite{delgado2023participatory, ram2020review,sorensen2023value,inie2023designing}. 
We engage with these stakeholder groups not to ``solve'' issues with NLP tools but to unearth concerns and themes \citep{sloane2022participation}.
We also make use of participatory design in the formation of our guiding principles; uniquely, we survey stakeholders connected to a specific healthcare topic to better illuminate tensions and priorities for NLP usage for healthcare.

\subsection{Maternal Health: Vulnerabilities and NLP Applications} 
\label{subsection:related-work-maternal-health}

The U.S. is experiencing an urgent and worsening maternal mortality crisis. 
Rates of pregnancy-related deaths and complications have increased over the last thirty years in the U.S., with particularly high rates among Hispanic and Black birthing people and non-U.S. citizens \citep{berg2010pregnancy,Creanga2014MaternalMA,Creanga2015-on,MacDorman2016-sf,Martin2017-nothing}. 
For example, Black women are three times more likely to die in childbirth than white women \citep{Creanga2015-on,Martin2017-nothing}.
Most frustratingly, 46\% of maternal deaths of Black women and 33\% of maternal deaths of white women are estimated as preventable \citep{berg2005preventability}, but measurements of these problems are challenging \citep{declercq2023prior}. 
Additional dangers face birthing people, such as postpartum depression \citep{Stewart2016postpartum}, and urgently need supportive solutions.

Researchers and healthcare practitioners have sometimes turned to NLP methods to try to address various challenges in the maternal health space.
For example, prior research has used NLP methods to study the prevalence of adverse outcomes via social media data \citep{lee-etal-2021-classification}, extract lactation information from drug labels \citep{goodrum-etal-2019-extraction}, predict high-risk pregnancies from electronic health records (EHR) \citep{knowles2014high}, examine how news media discuss pregnancy and exposure to isotretinoin \citep{mohammadhassanzadeh2020using}, and assist in analysis of phone interviews about breastfeeding \citep{valdeolivar2023towards}.
In our review of this prior work in NLP, we observe the following patterns: (i) many studies attempt to surveil and predict psychological states of the birthing person, e.g., predicting postpartum depression, and (ii) most studies use either social media or EHR data, with a bias in studies published at NLP venues towards social media data produced by birthing people.

\section{Methods}

\begin{figure}[H]
    \centering
    \includegraphics[width=0.9\textwidth]{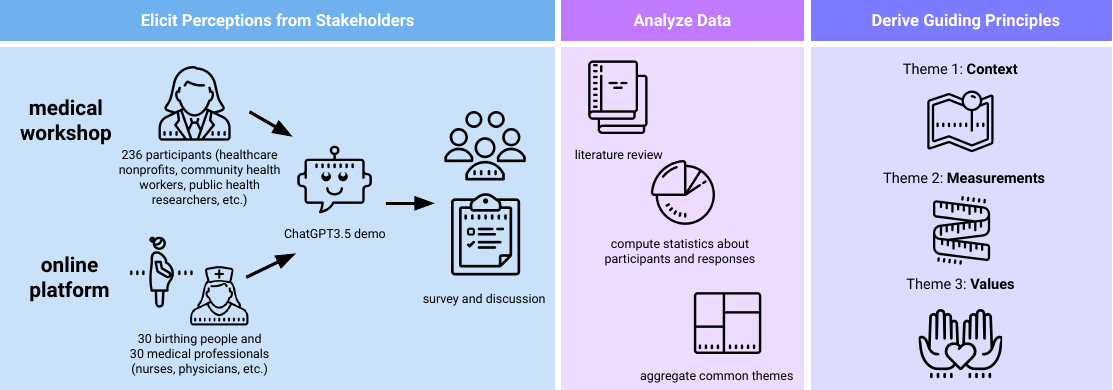}
    \caption{Study overview, including the three participant cohorts, the chatbot demonstration, and the surveys and discussions followed by analysis and design of the guiding principles.
    }
    \label{fig:overview}
\end{figure}

In Figure \ref{fig:overview}, we provide an overview of our methods.
Participants first engage with a structured interface to query an LLM-based chatbot.
After participants have interacted with this demonstration, they move to discussions (either virtual breakout discussion sessions led by volunteer moderators or independent written comments), and a survey, in which we asked participants to express their perceptions of NLP tools for maternal healthcare.
Our study was approved by the institutional review board (IRB) at the Allen Institute for AI.

\subsection{Participant Cohorts}

We ran this study with three participant cohorts.
For full demographic descriptions of these cohorts, see Table \ref{table:participants} in \S\ref{section:survey-results}.

\subsubsection{Healthcare workers and birthing people}
Using the Prolific\footnote{\url{https://www.prolific.com/}} online survey platform, we recruited one cohort of birthing people (gave birth in the last five years) and one cohort of healthcare workers (have ever worked in any healthcare profession).
We required that all participants be over 18 years old and be located in the U.S.
Prolific workers were paid an average of \$15/hour.

\subsubsection{Workshop participants}

For the third cohort, we led a live, virtual session with a group of participants who first received training about LLMs and their risks.
Our session was held at the end of a virtual workshop hosted by a U.S.-based medical nonprofit.
The workshop was themed around NLP and maternal health equity, and in sessions preceding ours, participants were introduced to the basics of NLP, heard research talks about applications of NLP to maternal health, and learned about biases and ethical challenges in NLP tools.
The workshop was open to the public, and participants were solicited through the nonprofit's email listserv, social media, and word of mouth.
Most but not all of these participants also worked in healthcare, though compared to our cohort from Prolific, these participants more often worked in community and research roles rather than as clinicians (see Table \ref{table:participants}).

\subsection{Design of chatbot demonstration}

We built an interface wrapper (Figure~\ref{fig:demo}) around GPT-3.5 from OpenAI.\footnote{Compared to GPT-4, which was also accessible at the time of our study, GPT-3.5 is less expensive and has lower rate limits, which was especially important for the workshop setting, where 100+ people were interacting with the bot at any moment. Our goal was to demonstrate high performance for the bot rather than create a reproducible system; otherwise, an open model would have been preferable.} 
Our goal with this demonstration was to give participants a guided but unconstrained interaction with a chatbot, gathering their queries, structured feedback, and perceptions related to the chatbot and maternal health. 
For many of our participants (see Figure \ref{fig:ai-familiarity} in the Appendix), this was their first direct interaction with a modern LLM.
We included prominent warnings about (i) privacy and data collection and (ii) the importance of always seeking the advice of a healthcare professional (see Figure \ref{fig:demo} in Appendix \ref{appendix:chatbot-demonstration}).
We also provided examples of the kinds of maternal health questions one might ask a chatbot.

\subsection{Design of discussion sessions}

\subsubsection{Workshop participants}
Workshop participants were divided into eight virtual breakout rooms and participated in a 30-minute free-form discussion session, moderated by the authors and trained volunteers. 
Discussion sessions were recorded after obtaining consent from all participants. 
During the session, participants were shown the following discussion prompts:
\begin{enumerate}
    \item How was your experience with the chatbot? What stood out to you about the responses?
    \item What are your dream NLP tools for maternal health? What tools should never be built?
    \item Which maternal health stakeholders (birthing people, nurses, doulas, etc.) would benefit or be hurt by NLP tools?
    \item What principles should guide the use of NLP for maternal health? What should be the goals and guardrails?
\end{enumerate}

\noindent We collected video and audio recordings for four out of eight sessions as well as written comments (collected via Mentimeter\footnote{\url{https://www.mentimeter.com/}}) from five sessions, which we included in our analysis.

\subsubsection{Healthcare workers and birthing people}
Like the workshop participants, non-workshop participants first interacted with our chatbot demonstration and then wrote answers to our discussion questions as part of a followup survey.
Unlike the workshop participants, non-workshop participants did not participate in discussions with groups or with a moderator but instead completed the study independently.

\subsection{Survey design}

Our goals in the survey were to (i) elicit participants' general perceptions of NLP tools, (ii) learn about participants' information seeking goals, and (iii) build collaborative rankings of values that should guide NLP uses for maternal health.
Surveys given to each cohort were identical except that for birthing people, questions about healthcare careers were replaced with questions about whether the participant would like their healthcare team to use and/or disclose their use of AI tools.
See Appendix \ref{section:full-survey-questions} for the full set of survey questions.

Participants were first asked about their generalized trust then asked about their trust in healthcare providers; the format of these questions is drawn from \citet{baughan2023mixed}.
Participants were then asked about their familiarity with NLP tools like ChatGPT, as well as their general perceptions of AI; these questions were taken from a frequently reused set by \citep{nickell1986computer}.
Next, participants were asked to select five out of ten \textit{ethical values} that should guide the use of NLP for maternal health. 
The listed values were taken from \citet{jakesch2022different} in a study of variations in attitudes towards AI by different demographic groups. 
We provided definitions (also drawn from \citet{jakesch2022different}) and example healthcare applications of NLP systems.
Participants were asked about their information seeking needs (where and to whom they turn with maternal health questions) and asked their opinion about the effects of NLP tools on maternal health team members and on their own job (if applicable). 
Finally, participants answered a series of demographic questions, including specifying any professional role they had ever taken in maternal health (e.g., worked as a midwife).
\section{Results}

\subsection{Survey Results}
\label{section:survey-results}

\begin{figure}[t]
    \centering
        \includegraphics[width=0.9\textwidth]{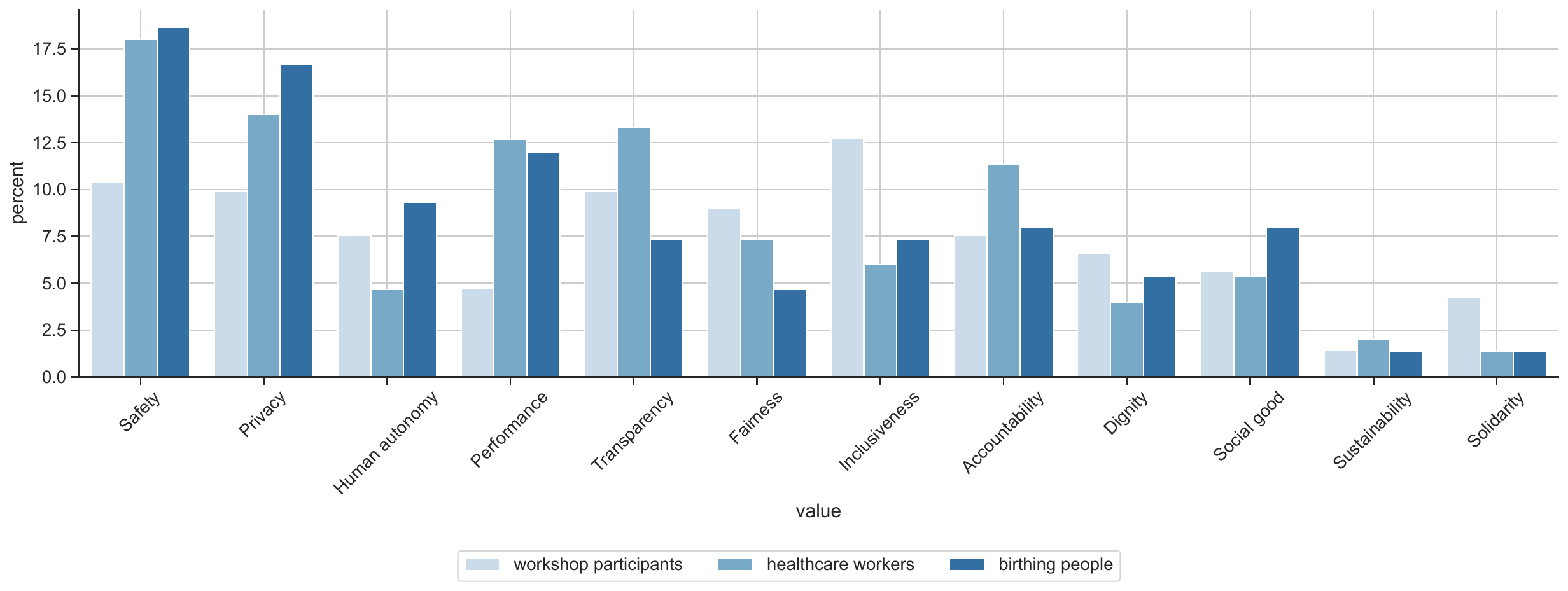}
    \caption{Frequencies of selected values. Each participant was asked to select five values from a list of 12 curated by \citet{jakesch2022different}. Participants were given definitions of the values, also drawn from \citet{jakesch2022different}. Birthing people and healthcare workers overall responded more similarly to each other than to the workshop participants, but this was not always the case (e.g., \textit{transparency}, \textit{human autonomy}).
    } 
    \label{fig:selected-values}
\end{figure}

We release our survey results publicly to support future research.\footnote{We remove demographic features for the workshop participants, as these participants interacted with one another and could potentially identify others' survey responses. Participants consented to the sharing of their responses. 
All survey data is available in supplementary materials and at a public Github repository.}
Figure \ref{fig:selected-values} shows how frequently each value was selected by each cohort when asked to ``select any five values ... that you think are most important for NLP systems for maternal health.''
Overall, \textit{safety}, \textit{privacy}, and \textit{performance} were selected more often by birthing people and healthcare workers in comparison to workshop participants, who were more likely to select \textit{inclusiveness}.
The birthing people and workshop participants were united in being more likely to select \textit{human autonomy} than the healthcare workers, and less likely than the healthcare workers to select \textit{transparency} and \textit{accountability}.
Overall, we do not find large differences in the value selections within cohorts across stratifications for generalized trust, trust in healthcare workers, or generalized AI perceptions (see Appendix \ref{app:survey-results-full}).

Table~\ref{table:participants} provides a summary of participant demographics and background. 
The workshop participants tended to represent the non-profit, community health, research, and governmental sectors in contrast to clinicians such as physicians and nurses that were more common in the Prolific cohort of healthcare workers. 
The workshop participants represented a more even distribution across racial/ethnic groups, but all cohorts had little representation in the East Asian, South Asian, and Southeast Asian groups.\footnote{We attempted to correct this balance on Prolific by supplementing our results with an additional cohort of East Asian, South Asian, and Southeast Asian participants, but we found that the Prolific survey platform included very few such participants who also met our other criteria (e.g., gave birth in the last five years).}
Overall, workshop participants reported less experience with AI and NLP than birthing people and healthcare workers recruited from Prolific (see Appendix \ref{app:survey-results-full}).

\begin{table*}[th!]
    \centering
    \footnotesize
    \begin{tabular}{@{}L{5.0cm}L{3.1cm}L{1.2cm}L{2.3cm}L{1.6cm}@{}}
    \toprule
        \textbf{Cohort} & \textbf{Race/Ethnicity} & \textbf{Age} & \textbf{Highest Education} & \textbf{Gender} \\ 
        \midrule \\ [-7pt]
        \multicolumn{5}{c}{\textit{Workshop Participants ($N=39$)}} \\ [1pt]
        \midrule \\ [-7pt]
        38\% community nonprofits \newline 27\% pop./public health research \newline 24\% comm. health/promotara \newline  24\% local/state public health \newline 19\% healthcare management/admin \newline  16\% healthcare services researcher \newline 11\% other perinatal healthcare provider \newline 11\% other non-healthcare perinatal support \newline 8\% doula \newline 8\% non-perinatal healthcare provider\newline  13.5\% \textit{all other groups} 
        & 41\% African-American/Black \newline 41\% White \newline 16\% Hispanic/Latino/a/x \newline 5\% South Asian \newline 19\% \textit{all other groups}
        & 35\% 35-44 \newline 30\% 25-34 \newline 19\% 45-54 \newline 11\% 55-64 \newline 5\% 65-74 
        & 38\% MS, MPH, etc. \newline 30\% PhD \newline 24\% BA, BS, etc. \newline 11\% \textit{all other groups}
        & 92\% women \newline 5\% men \newline 3\% no answer \newline 0\% non-binary
        \\
        \midrule \\ [-7pt]
        \multicolumn{5}{c}{\textit{Healthcare Workers ($N=30$)}}\\ [1pt]
        \midrule \\ [-7pt]
        20\% nurse \newline 17\% pharmacy \newline 10\% physician \newline 10\% medical tech \newline 10\% medical assistant/aide \newline 10\% research \newline 23\% \textit{all other groups} \newline 33\% have worked in maternal/perinatal healthcare
        & 57\% White \newline 23\% African-American/Black \newline 7\% East Asian \newline 7\% Southeast Asian \newline 9\% \textit{all other groups}
        & 33\% 35-44 \newline 30\% 25-34 \newline 10\% 18-24 \newline 3\% 65-74 \newline 3\% 55-64
        & 50\% BA, BS, etc. \newline 17\% MS, MPH, etc. \newline 17\% Trade School \newline 10\% MD, DO, etc. \newline 7\% Community College \newline 6\% \textit{all other groups}
        & 79\% women \newline 21\% men \newline 0\% non-binary \\
        \midrule \\ [-7pt]
        \multicolumn{5}{c}{\textit{Birthing People ($N=30$)}} \\ [1pt]
        \midrule \\ [-7pt]
        20\% have worked in healthcare \newline 7\% have worked in maternal/perinatal healthcare
        & 73\% White \newline 20\% Hispanic/Latino/a/x \newline 17\% African-American/Black \newline 12\% \textit{all other groups}
        & 53\% 25-34 \newline 37\% 35-44 \newline 10\% 65-74
        & 33\% BA, BS, etc. \newline 30\% High school or GED \newline 13\% Community College \newline 10\% MS, MPH, etc. \newline 10\% Trade School \newline 7\% PhD \newline 3\%  Prof. Degree 
        & 97\% women \newline 7\% non-binary \newline 0\% men
        \\
        \bottomrule \\
    \end{tabular}
    \caption{Demographic description of participant cohorts. Healthcare workers and birthing people were recruited from the Prolific platform while the workshop participants took part in multiple training and educational sessions to learn about LLMs, their applications to maternal health, and their risks. Work category, race/ethnicity, highest education, and gender all allowed multiple selections, so the percents for these categories might not sum to 100\%.
    }
    \label{table:participants}
\end{table*}

\subsection{Thematic Analysis of Discussion Sessions}

We identified themes and concepts frequently brought up by study participants in workshop breakout discussions, written comments, free-text survey answers, and chatbot queries. 
We first performed open coding on responses from each source to identify relevant concepts \citep{charmaz2014constructing}, then axial coding to group these concepts under broad themes \citep{Corbin1990GroundedTR}. 
The authors collectively discussed these themes and iterated several times before synthesizing them into the final set of three themes and nine guiding principles presented in \S\ref{ssec:principles}.
We summarize these principles in the following section and include anonymized quotes from our study participants to highlight stakeholder pain points and further support the inclusion of each principle. 
Additionally, we provide actionable recommendations to NLP researchers and practitioners based on our own experience and expertise as NLP researchers and healthcare workers.
Our thematic analyses also raised several additional points, which we discuss in \S\ref{section:discussion}. 

\section{Guiding Principles}
\label{ssec:principles}

Drawing from the discussion themes, our survey results, and related work (both applied work and high-level ethics guidelines like those by \citet{chen2021ethical} and \citet{dignazio2023data}), we develop guiding principles for the responsible use of NLP in maternal health.

We present these principles grouped under three themes: \textit{context}, \textit{measurements}, and \textit{values}. 
\textit{Context} principles ask practitioners to incorporate the fundamentals and history of maternal health in their applications, \textit{measurement} principles discuss what to optimize for and how to evaluate, and \textit{value} principles address how practitioners should situate user voices and data relevant to their systems.

We summarize each principle below and highlight in participants' own words their importance. 
In quotes, workshop participants are attributed as W1-10, birthing people as B1-11, and healthcare workers as H1-15.

\subsection{Theme 1: Context}

\noindent \textbf{Be aware of power dynamics in the care team.}
~~\citet{dignazio2023data} have argued that \textit{power} should be a central concern of feminist data science, and maternal healthcare has a long and fraught history of shifting power dynamics in the care team \citep{benjamin2022better}. 
These historical shifts, e.g., male OB/GYNs replacing midwives in the early 20th century \citep{Brodsky2008WhereHA, Loudon2008GeneralPA}, have led to the marginalization of some maternal health workers 
and both good and bad consequences for birthing people. 
NLP practitioners should know this history and work to improve rather than exacerbate existing hierarchies, being mindful of the impact of tools and automation on midwives, doulas, and other care workers whose placement is already precarious. 
For example, one of the healthcare workers in our study mentioned that \textit{``an AI chatbot can eliminate the need for a Doula''} (H1), highlighting the vulnerabilities of certain workers. 
One of the birthing people framed this as a harm rather than an opportunity: \textit{``If mothers relied too heavily on AI instead of seeking professional help then the nurses and doulas may see fewer people seeking care''} (B11).

\begin{tcolorbox}
\hspace{-2mm}\textit{Recommended actions for practitioners:}
\begin{itemize}[noitemsep,topsep=0pt,leftmargin=5pt]
    \item  Understand historical power shifts in the care team. Consider carefully who you expect to use your tool and who will be impacted (or potentially replaced) by your tool. Rather than designing a chatbot as a replacement for a core member of the care team, consider designing supportive tools that can improve the care team's collaboration.
    \item  Consider these questions: Whose language and data are represented in the training data of the tool? Who will receive predictions and advice? Who will have access to and control? Are you concentrating more power within a single care team role?
\end{itemize}
\end{tcolorbox}

\vspace{2mm}
\noindent \textbf{Know the politics and implications of your measurements.}
~~Using NLP tools for data collection can have unintended consequences that can impact the \textit{safety} and \textit{agency} of birthing people.
For example, (i) overemphasis on data collection can draw resources away from building solutions \citep{dignazio2023data}, 
(ii) reproductive health is politicized, and measurements may be used as evidence in support of unexpected agendas, 
and (iii) focusing only on measuring problems can contribute to ``deficit narratives'' that blame communities for their own challenges \citep{dignazio2023data}. 
In our literature review (\S\ref{section:background}), we found that much prior NLP research for maternal health has focused on data collection and surveillance of birthing people, and participants pushed back against this pattern:
\textit{``It's key that the AI is actually able to provide solutions to problems that can be fixed, and not just simply acknowledging them''} (H4). 
Financial motivations also muddy these decisions, with the healthcare industry often focused on reducing costs \citep{obermeyer2019dissecting}; as one participant put it, \textit{``I get worried about what'll happen when insurance companies think there’s cost savings to using these tools...they can cut corners, have more profit...given the incentives in ... the healthcare system''} (W1).

\begin{tcolorbox}
\hspace{-2mm}\textit{Recommended actions for practitioners:}
\begin{itemize}[noitemsep,topsep=0pt,leftmargin=5pt]
    \item Clarify your research and/or impact goals before collecting data. 
    \item Consider the current allocation of resources and where help is most needed when choosing research questions. While additional data collection can confirm well-known problems, it can also draw resources away from other important problems.
    \item Think through possible narratives involved in public portrayals of your work, and invest time in framing your NLP and especially LLM outputs for better public understanding of risks (e.g., hallucinations) involved in these technologies.
    \item Know that your measurements and predictions could contain information that could be used to incriminate the birthing person in some places and situations. Similarly, your data and measurement methods could be used for unintended purposes, like personalized advertising and the setting of insurance rates. 
\end{itemize}
\end{tcolorbox}

\vspace{2mm}
\noindent \textbf{Learn from maternal health traditions and communities.}
~~Thousands of years of cultural traditions, healthcare practices, and tool development related to maternal health already exist \citep{goode2017african}, and more recently, grassroots communities on the internet have formed to help birthing people prepare for and work through their challenges and experiences  \citep{antoniak2019narrative,andalibi2021sensemaking}. 
\textit{``People have the mindset that this needs to actually replace things or that this needs to be the way in the future of addressing maternal health...diminishing all the work that's been done in the past and the positives of having people that truly understand this work.''} (W2) 
Learning from traditions and communities can help avoid repeating mistakes made by others. 
\textit{``It [(AI)] should follow the same guidelines that medical professionals do: `Do no harm.'\thinspace''} (H6)

\begin{tcolorbox}
\hspace{-2mm}\textit{Recommended actions for practitioners:}
\begin{itemize}[noitemsep,topsep=0pt,leftmargin=5pt]
    \item Combining NLP expertise with an interdisciplinary team can help you avoid reinventing the wheel.
    \item Consider how NLP tools can support and learn from storytelling, which is an important language-based way for communities and generations to collectively teach and learn about maternal health, as well as a way for birthing people to process their experiences \citep{kay2017engaging,carson2017narrative}. 
    \item Consider how to sample training datasets without dismissing the care-seeker's perspective; weigh the risks of possible misinformation against the value of personal experiences within your specific healthcare setting. 
\end{itemize}
\end{tcolorbox}

\subsection{Theme 2: Measurements}

\vspace{2mm}
\noindent \textbf{Optimize for outcomes that support the whole person.}
\textit{``It’s not just about the outcome, right? It’s about the whole experience''} (W3). 
Most NLP research is centered on structured tasks with defined and accessible outcomes, e.g., predicting postpartum depression.
Such tasks are important, but so are other outcomes, like achieving a positive birthing experience \citep{Karlstrm2015TheMO} or breastfeeding without stress \citep{Spannhake2021ItIA}. 
As one participant says, \textit{``it would be great if AI could tell everything from conception to birth''} (B10). 
Taking a more expansive view of the tasks and optimizations that could benefit maternal health stakeholders can open the door to other opportunities. 
By ``expansive view,'' we do not mean that a single model must be capable of all tasks (as this may in fact be a weakness of current LLMs whose purported generalizability can lead to a lack of clear evaluation \citep{raji2021ai}). 
Rather, we invite NLP researchers and practitioners to be open-minded about tasks they can address with their NLP work.
\textit{``A dream AI tool should have the ability to address and develop solutions for mothers who are struggling to cope with the demands of pregnancy and overall maternal health''} (H4).

\begin{tcolorbox}
\hspace{-2mm}\textit{Recommended actions for practitioners:}
\begin{itemize}[noitemsep,topsep=0pt,leftmargin=5pt]
    \item  Curate and create text datasets with novel annotations, valuing this data work \citep{sambasivan2021everyone}.
    \item  Use social media and birthing person interviews to get a holistic overview of what outcomes birthing people care about \citep{andalibi2021sensemaking}.
    \item  Run long-term user studies to support a broader range of outcomes of interest.
    \item Consider including subjective and/or qualitative measurements.
\end{itemize}
\end{tcolorbox}

\vspace{2mm}
\noindent \textbf{Protect all groups of birthing people.}
~~The healthcare system treats people differently on the basis of race, class, gender, etc. \citep{Smedley2003UnequalTC, Zavala2020CancerHD, WebbHooper2020COVID19AR}. 
NLP models can mirror biases found in training datasets \citep{Caliskan2016SemanticsDA, steed-etal-2022-upstream}, and even worse, they can over-represent and exacerbate the impacts of those biases \citep{Zhao2017MenAL, Hall2022ASS}. 
\textit{``Everything that has the potential to benefit also has the potential to hurt''} (W4). 
When evaluating NLP systems, one should consider both disparities in impact (health outcomes for different groups) and disparities in treatment (whether people with the same complaint receive different treatment). 
Tools should \textit{``allow providers to improve their own cultural competency or race-consciousness''} (W5), and empower 
\textit{``those lacking social and financial support,''} such as young birthing people (H14).

\begin{tcolorbox}
\hspace{-2mm}\textit{Recommended actions for practitioners:}
\begin{itemize}[noitemsep,topsep=0pt,leftmargin=5pt]
    \item Recruit birthing people from diverse groups to participate and/or give feedback on your NLP study/tool.
    \item Consider the role of personalization in your NLP/LLM-based system, considering both disparate impacts and disparate treatments. Sometimes people with the same complaint are not given the same treatment (e.g., pain management disparities \citep{Samulowitz2018BraveMA}). But in other cases, applying the same treatment, without regard for individual circumstances, is also an inequity.
    \item When building LLM-based tools that communicate with birthing people (e.g., chatbots), evaluate across a diversity of vocabularies, accents, and languages \citep{mehandru2022reliable}. 
    \item Beware counterfactuals as a methodological bandaid \citep{kasirzadeh2021use}. Measuring and addressing disparities are important, but simply switching tokens indicating race or gender (e.g., replacing all mentions of \textit{she} with \textit{he}) ignores society's multi-dimensional impacts on people living with those identities, 
    which can lead to correlated differences in text datasets. 
\end{itemize}
\end{tcolorbox}

\vspace{2mm}
\noindent \textbf{Hold onto the human: empathy, emotion, relationships, complexity.}
~~Emotion, empathy, and storytelling are important components of healthcare \citep{charon2001narrative}, and NLP tools should account for such human elements, remembering that each birth is unique and each birthing person has a unique set of circumstances, experiences, and preferences. 
\textit{``\textit{Human connection is important and should be emphasized so that better quality service can be provided}''} (W6); \textit{``\textit{your own judgment and/or human compassion components, wisdom, experience [are a] part of the care}''} (W7). 
Working with language data, as opposed to structured datasets, opens doorways to subjective tasks such as measuring empowerment \citep{njoo-etal-2023-talkup}, agency \citep{sap-etal-2017-connotation}, and sentiment \citep{ali-etal-2013-hear}.
\textit{``Birth is so complicated though that I don’t see women ever replacing medical care with AI advice''} (B11). \\ [-3mm]

\begin{tcolorbox}
\hspace{-2mm}\textit{Recommended actions for practitioners:}
\begin{itemize}[noitemsep,topsep=0pt,leftmargin=5pt]
    \item Look for outliers and value what they can teach you, rather than removing them. 
    \item A one-size-fits-all approach is probably not appropriate; prioritize personalization and/or allow the tool to be tailored by the individual birthing person or healthcare worker.
    \item Subjective language tasks like measuring framing \citep{card-etal-2015-media} can illuminate aspects of healthcare experiences overlooked by non-text-based methods.
    \item Include qualitative methods in your study design, complementing NLP models with interviews, grounded theory \citep{Corbin1990GroundedTR}, and methods from the social sciences.
\end{itemize}
\end{tcolorbox}

\subsection{Theme 3: Values}

\noindent \textbf{Include the voices of those seeking care.}
~~The voices of the most vulnerable stakeholder group, birthing people, should be included in the design and development of NLP tools. 
Center the principle of ``nothing about us without us,'' most recently popularized by disability activists \citep{Charlton1998NothingAU}. 
While it may not always be possible to include birthing people on a research or production team, it is important to integrate their perspective throughout design and deployment processes through surveys and user studies. 
As one participant put it, \textit{``there needs to be community input, there needs to be representation in creating these tools''} (W8). 
At the same time, practitioners should be careful to engage fully with these voices, rather than using them as an ethical veneer on otherwise exploitative tools \citep{sloane2022participation}.

\begin{tcolorbox}
\hspace{-2mm}\textit{Recommended actions for practitioners:}
\begin{itemize}[noitemsep,topsep=0pt,leftmargin=5pt]
    \item  Where possible, include birthing people and healthcare professionals in your research team.
    \item  Include surveys and user studies throughout your research and design process.
    \item  Incorporate literature written by birthing people, and learn from related work about the birthing experience and birthing people’s perspectives and needs.
    \item  Be gender inclusive: do not automatically predict the gender of study subjects \citep{keyes2017stop,larson-2017-gender}, and remember to include the concerns of trans birthing people in your study design. Use gender inclusive language and follow the HCI Guidelines for Gender Equity and Inclusivity \citep{scheuerman2020hci}.  
\end{itemize}
\end{tcolorbox}

\vspace{2mm}
\noindent \textbf{Always center the agency and autonomy of the birthing person.}
~~Maternal healthcare has an unfortunate history of abuse and disregard for the birthing person’s agency (see \S\ref{subsection:related-work-maternal-health}). 
NLP practitioners, even when studying topics like misinformation, should not work from a perspective of skepticism about the birthing person's ability to make decisions for themselves.
How NLP tools are being used to make decisions in maternal healthcare
should be disclosed to the birthing person, such that \textit{``the person using it knows what kind of information or advice it can and cannot give''} (B2). 
Correspondingly, some tools should not be built \citep{bender2021dangers}, such as anything \textit{``infringing on the rights of the mom or baby''} (B3).

\begin{tcolorbox}
\hspace{-2mm}\textit{Recommended actions for practitioners:}
\begin{itemize}[noitemsep,topsep=0pt,leftmargin=5pt]
    \item Explore the construction of NLP tools that increase the agency available to the birthing person rather than making decisions for them. For example, NLP tools might be used to provide birthing people with additional decision points, resources, explanations, descriptions of what to expect, or assistance for communicating with healthcare providers.
    \item If NLP tools are used to make or assist in clinical decisions, this should be disclosed to birthing people, and where possible, providers should obtain direct consent from birthing people. 
\end{itemize}
\end{tcolorbox}

\vspace{2mm}
\noindent \textbf{Respect and support your data sources.}
~~Generations of birthing people have passed down oral stories, written books, and created online content about pregnancy, labor, and the postpartum period. 
NLP studies and tools benefit from this knowledge and data and should give credit and be designed to avoid supplanting systems of support that are already thriving. 
As several workshop participants observed, ChatGPT and other similar chatbots \textit{``don’t really note their references of where they’re getting the information from''} (W9).
\textit{``The principle that should guide these tools is to... have transparency for its sourcing of data''} (H2).

\begin{tcolorbox}
\hspace{-2mm}\textit{Recommended actions for practitioners:}
\begin{itemize}[noitemsep,topsep=0pt,leftmargin=5pt]
    \item Give credit to the data sources. Use proper attribution, as this not only respects people’s work but also builds trust from your users and supports auditing of your system.
    \item Maintain privacy. Collect only necessary data and store the data securely. Avoid perpetuating the over-surveillance of birthing people.
    \item Beware the ``paradox of reuse'' \citep{McMahon2017TheSI} in which the creation of an automated tool removes incentives for people to continue creating the training texts that power that LLMs. Encourage users to re-engage with the data source (e.g., by posting their own experiences to an online community).
\end{itemize}
\end{tcolorbox}
\section{Discussion}
\label{section:discussion}

\subsection{Perceptions of risks and benefits of LLMs} 

Participants reported largely positive attitudes towards NLP tools with important caveats.
For example, workshop participants expressed high hopes about the development and adoption of NLP tools in healthcare but unanimously agreed that NLP tools for maternal health should always be designed to be assistive rather than autonomous (\emph{holding on to the human}).
Settings in which NLP tools were viewed to be particularly useful to healthcare providers included: (i) reducing administrative burden, including better/faster communication and coordination of care across medical departments, and (ii) improving medical education and training, including assisting staff in developing better cultural sensitivity. 
On the other hand, tools for decision-making settings were viewed negatively. 
These results contrast with current trends in NLP research (\S\ref{section:background}), which tend towards surveillance of birthing people and tasks like risk prediction.

Birthing people viewed NLP tools very positively, especially for providing information and recommendations (\emph{centering their agency and autonomy}). 
For example, participants wrote that they wished they had the chatbot during their pregnancy: 
\textit{``I wish it [(AI)] had been around when my son was a newborn so I could interact with it during late night feedings.  One, to give me something to do, and two, to make me feel like I wasn't alone''} (B12).
In particular, birthing people emphasized how comforting it would be to have a fast and convenient resource to assuage fears about whether what they were experiencing was normal: \textit{``Just to have something there to ask questions to when I am not sure as to what is happening or when I need a quick answer''} (B9). 
Some participants cautioned that such tools should provide additional context to help patients understand whether the suggestions or advice are applicable to their specific situation due to differing levels of medical understanding.

These positive impressions need to be weighed carefully against research demonstrating potential harms.
On one hand, both birthing people and clinicians are desperately in need of support, as e.g., postpartum depression rates \citep{Stewart2016postpartum} and clinician burnout rates \citep{dzau2018care} demonstrate.
Challenges around healthcare costs, inaccessibility, and community distrust of the medical establishment \citep{cuevas2016african} support the positive views of our participants towards NLP and LLM-based tools.
But on the other hand, the current capabilities of these tools are limited and their vulnerabilities well-documented \citep{weidinger2021ethical}, and for some challenges, non-technical solutions already exist whose implementation might be further delayed by allocating resources to NLP technologies.
Researchers and practitioners should be ready to abandon NLP technologies for maternal health if there are not clear benefits.

\subsection{LLMs as part of a larger ecosystem to meet information needs}

In their free-text responses, about half of the birthing people and two healthcare workers compared their experience with the chatbot to their past experiences with internet search engines.
Most framed the comparison favorably, writing \textit{``Often times people will google questions and try to sift through all the search results to find the applicable information. AI could make that a much more efficient process''} (H9) and \textit{``It would be nice to be able to type in worries and fears to an AI bot and get accurate answers instead of going down rabbit trails on search engines that leave you more concerned''} (B11). 
On the other hand, one participant noted dangers to LLMs similar to known dangers of popular healthcare websites: \textit{``People already diagnose themselves on WebMD. Providing more tools can be dangerous''} (B4).
These participants viewed LLMs as one more tool in an existing information ecosystem, exacerbating or ameliorating many of the same risks.
This view is complicated by recent work by \citet{shah2023envisioning} that emphasizes the risks of using LLMs as replacements for traditional search interfaces, both to their users (via hallucinations) and to entire information ecosystems that can be polluted by incorrect and ungrounded outputs of LLMs; this work calls for slower research that prioritizes answering fundamental questions about evaluation, costs, etc. before deploying LLMs as information retrieval systems.

\subsection{Comparison to prior ethics guidelines} 

Multiple guidelines for machine learning applied to healthcare applications already exist \citep{chen2021ethical,mccradden2023whats,Wiens2019DoNH,sendak2020human,petti2023ethical}; what does our study add to these?
(i) Our study is the first to focus on maternal health; this is a critical area with infamous historical abuses that warrants its own ethical studies.
Our work is also novel in (ii) our focus on new NLP technologies like LLMs and (iii) grounding our guiding principles in perceptions from multiple stakeholder groups.
Finally, (iv) the principles we uncover are distinct from those in prior work.
While some of the themes overlap with prior work (e.g., problem formulation \citep{sendak2020human}), many of our themes (e.g., power dynamics in the care team, political implications of measurements) have not been raised in prior guidelines, which have often been focused on machine learning vulnerabilities rather than the social systems surrounding these technologies.
Compared to the two studies closest to ours by \citet{sendak2020human} and \citet{petti2023ethical}, which focus on ethical guidelines for sepsis and Alzheimer's disease, respectively, our focus on maternal health uncovers principles specific to NLP and LLMs and emphasizes communities of birthing people and other care-seekers: their knowledge and tool development, their autonomy, and their historical and ongoing vulnerabilities.

\subsection{Generalization to other healthcare settings}

Maternal health served as a representative health case study through which we could apply a focused lens to the use of NLP and LLMs for healthcare.
Maternal health is a pressure cooker of modern biases, historical injustices, misinformation, care team imbalances, and other challenges that appear in many other healthcare contexts.
Related healthcare settings include resource shortages in primary care \cite{bodenheimer2013primary}, individuals seeking health information to diagnose symptoms~\cite{ayers2023comparing}, and racial considerations in disease severity models~\cite{omiye2023large}. 
In each case, our guiding principles can help practitioners think through whether and how to incorporate NLP tools into new healthcare settings.

\subsection{Limitations}
\label{subsection-limitations}

Our survey respondents are not representative of all birthing people worldwide. 
The birthing people in our study were from the U.S., spoke English, and lack representation among Asian and Pacific Islander groups. 
We used ChatGPT-3.5 for the demonstration, since its rate limits fit the constraints of the live workshop, though the model is regularly updated and results may not be replicable.
We highlight the need for longitudinal studies of LLM usage in the healthcare space, to illuminate when to use LLMs in the clinical tool development process and also the compounding effects of AI and LLMs on birthing people over time.

Each of our participant cohorts had a unique selection of values, forming individual priority ``signatures'' (Figure \ref{fig:selected-values}). 
Differences in results between our participant cohorts are likely driven by at least three factors, though we cannot make any conclusive causal claims: (i) the groups represent different stakeholders;
(ii) the workshop participants were primed by their multi-hour exposure to lectures about NLP biases; and (iii) the workshop participants opted into the workshop, which was themed around ``maternal health equity,'' while the Prolific participants were solicited online and paid for their work.
We also observed a notable difference in prior AI usage and general perceptions of AI between the workshop participants and the two groups of healthcare workers and birthing people recruited from Prolific. 
Workshop participants on average were much less familiar with AI and held less positive perceptions of AI (Figures~\ref{fig:ai-familiarity} and~\ref{fig:ai-perception} in the Appendix).

It is important to engage with multiple groups in different settings and integrating their viewpoints, not by enforcing agreement but by surfacing the inherent tensions and different viewpoints that exist in the perceptions of these technologies. 

\section{Conclusion}
In consultation with healthcare professionals, birthing people, and other stakeholders, we developed a set of guiding principles for the use of NLP and LLMs in maternal healthcare. 
This work serves as a guide to researchers and practitioners broadly on how to engage with affected peoples in healthcare contexts. 
We hope that researchers and clinicians will work with the people most affected---the people seeking care---and will read and follow the principles derived by our methods. 
Machine learning researchers should also explore how their tools can actively broadcast these principles or how model development pipelines can be adapted to better meet the needs of all stakeholders.


\bibliographystyle{ACM-Reference-Format}
\bibliography{bibliography}

\appendix

\section{Extended Survey Results}
\label{app:survey-results-full}

\begin{figure}[H]
    \centering
    \includegraphics[width=0.8\textwidth]{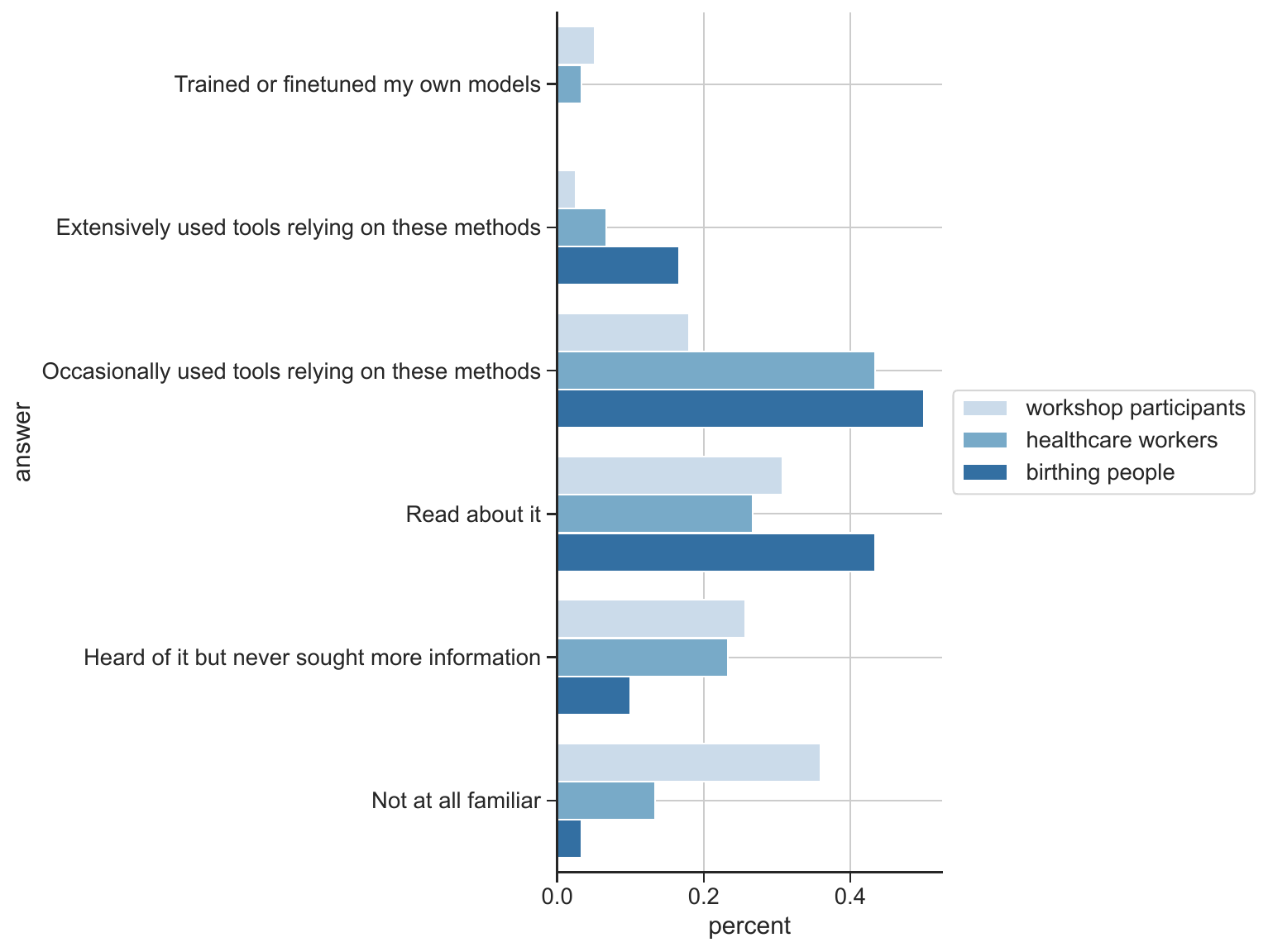}
    \caption{Answers to the survey question \textit{[Before this workshop,] How familiar were/are you with NLP, machine learning, and/or AI?} Overall, the workshop participants were less familiar with these topics than the healthcare workers and birthing people who were recruited via the Prolific platform.
    \newline
    \newline
    \textit{Alt Text: A bar plot showing the answers to the survey question about familiarity with NLP/ML/AI, with responses broken apart by participant cohort.}
    }
    \label{fig:ai-familiarity}
\end{figure}

\begin{figure}[H]
    \centering
    \includegraphics[width=0.4\textwidth]{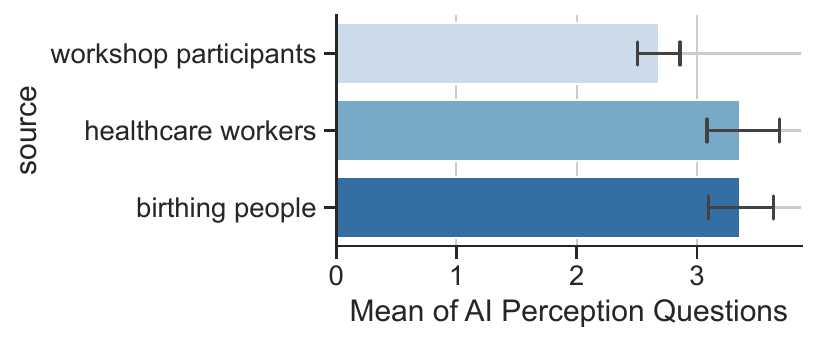}
    \caption{Answers to the four survey questions about general perceptions of AI. Overall, the workshop participants less frequently reported positive perceptions than the healthcare workers and birthing people who were recruited via the Prolific platform.
    \newline
    \newline
    \textit{Alt Text: A bar plot showing the averaged answers to the survey questions about general perceptions of AI, with responses broken apart by participant cohort.}}
    \label{fig:ai-perception}
\end{figure}

\begin{figure}[H]
    \centering
    \includegraphics[width=0.4\textwidth]{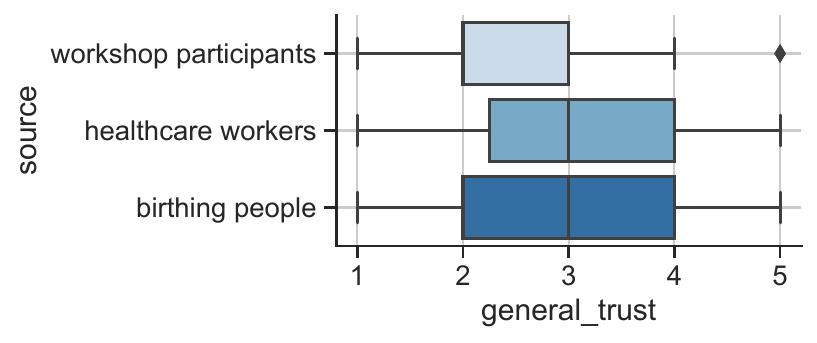}
    \includegraphics[width=0.4\textwidth]{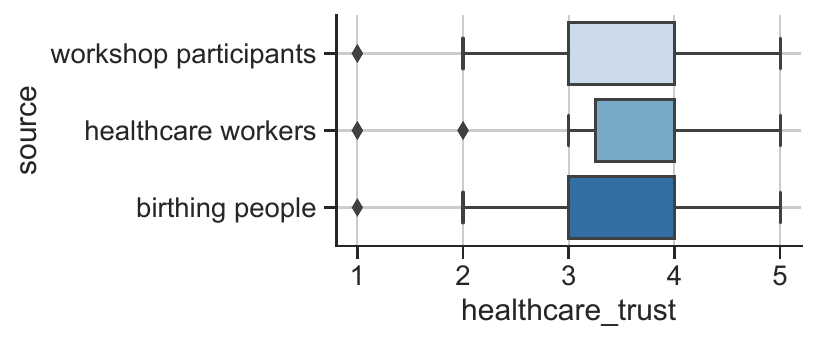}
    \caption{Answers to the two survey questions about trust. Overall, the healthcare workers were more trusting of healthcare providers.
    \newline
    \newline
    \textit{Alt Text: Two box plots showing the answers to the survey questions about trust, with responses broken apart by participant cohort.}}
    \label{fig:trust}
\end{figure}

\begin{figure}[H]
    \centering
    \includegraphics[width=0.6\textwidth]{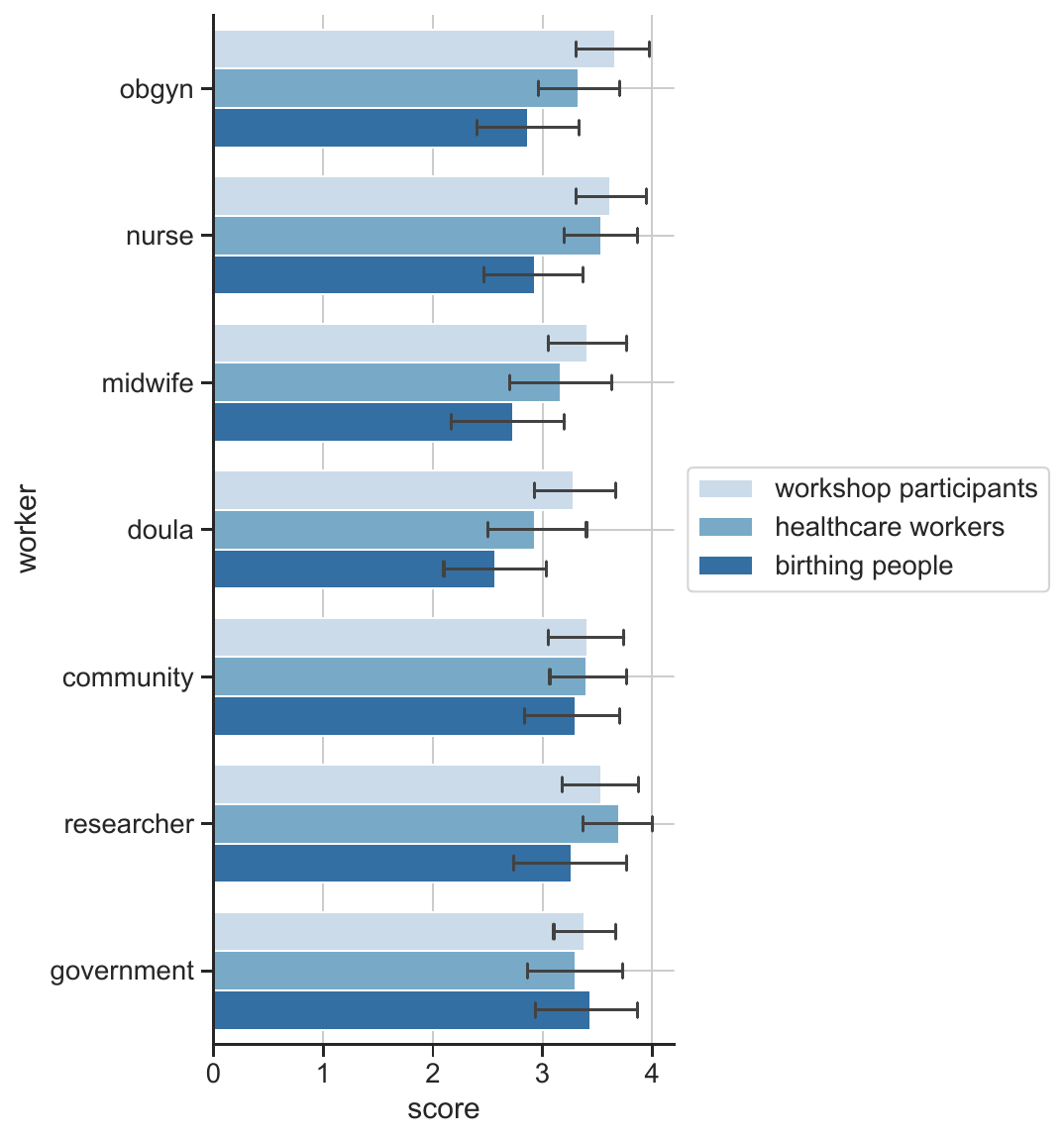}
    \caption{Answers to the survey questions about who would benefit from an AI/NLP chatbot in their work, with the 5-point Likert scale score shown on the x-axis.
    \newline
    \newline
    \textit{Alt Text: A bar plot showing the answers to the survey question about who would benefit from an AI/NLP chatbot, with responses broken apart by participant cohort.}}
    \label{fig:worker-benefits}
\end{figure}

\begin{figure}[H]
    \centering
    \includegraphics[width=0.9\textwidth]{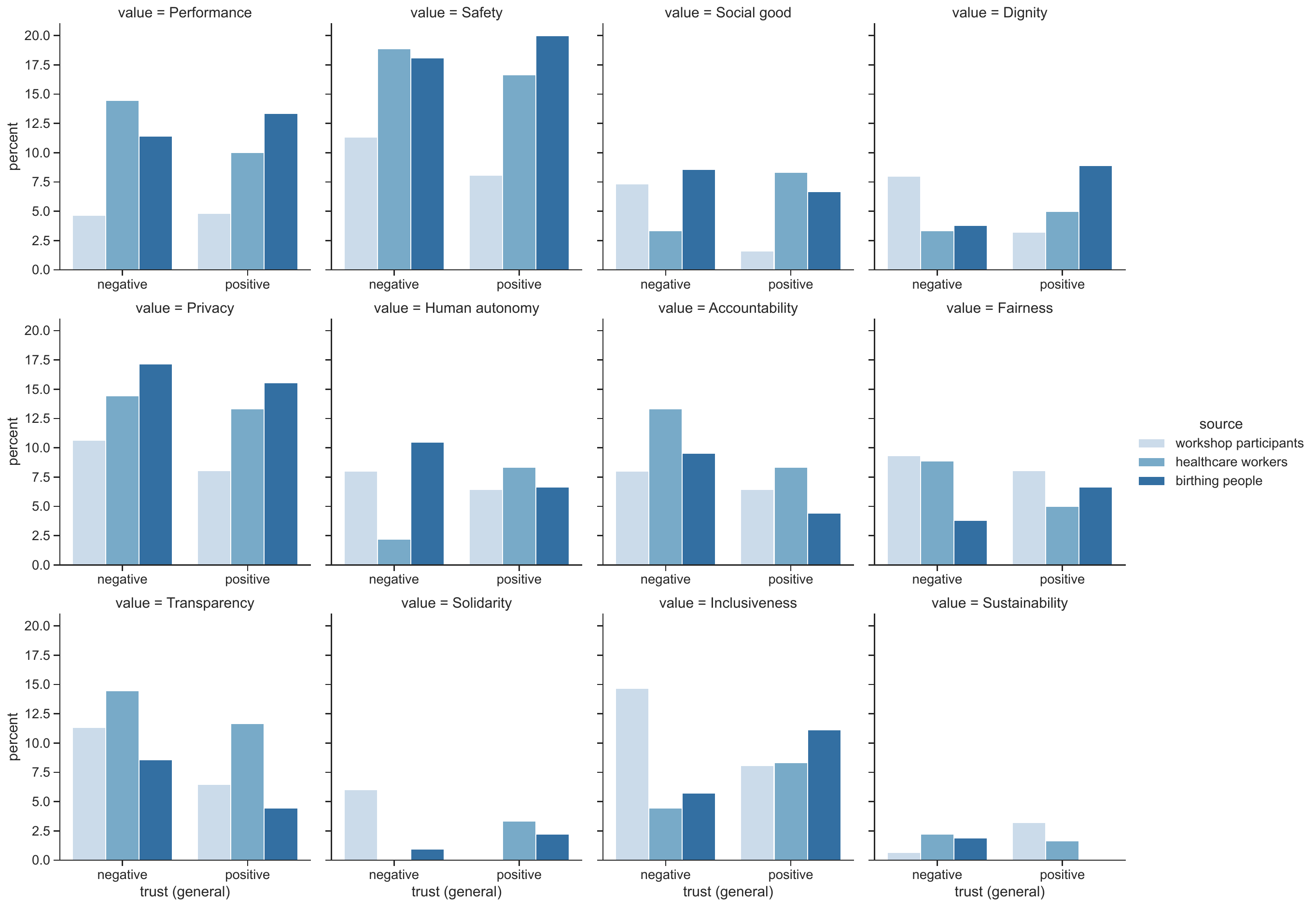}
    \caption{Selection of values broken apart by cohort and \textbf{general} trust. \textit{Positive} indicates scores $>3$.
    \newline
    \newline
    \textit{Alt Text: A set of bar plots showing the selection of values, with responses broken apart by participant cohort and generalized trust.}}
\end{figure}

\begin{figure}[H]
    \centering
    \includegraphics[width=0.9\textwidth]{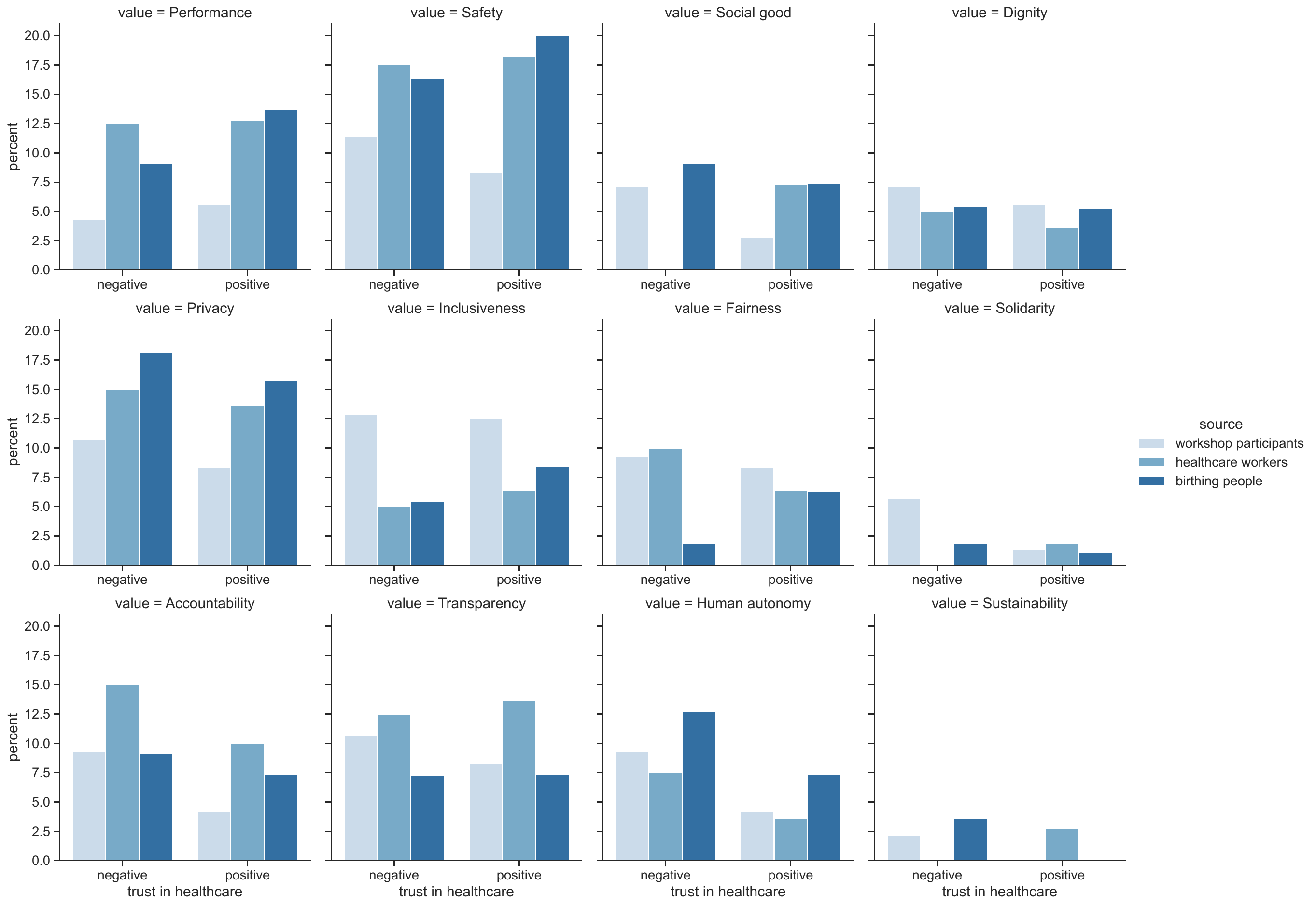}
    \caption{Selection of values broken apart by cohort and trust in \textbf{healthcare providers}. \textit{Positive} indicates scores $>3$.
    \newline
    \newline
    \textit{Alt Text: A set of bar plots showing the selection of values, with responses broken apart by participant cohort and trust in healthcare providers.}}
\end{figure}

\begin{figure}[H]
    \centering
    \includegraphics[width=0.9\textwidth]{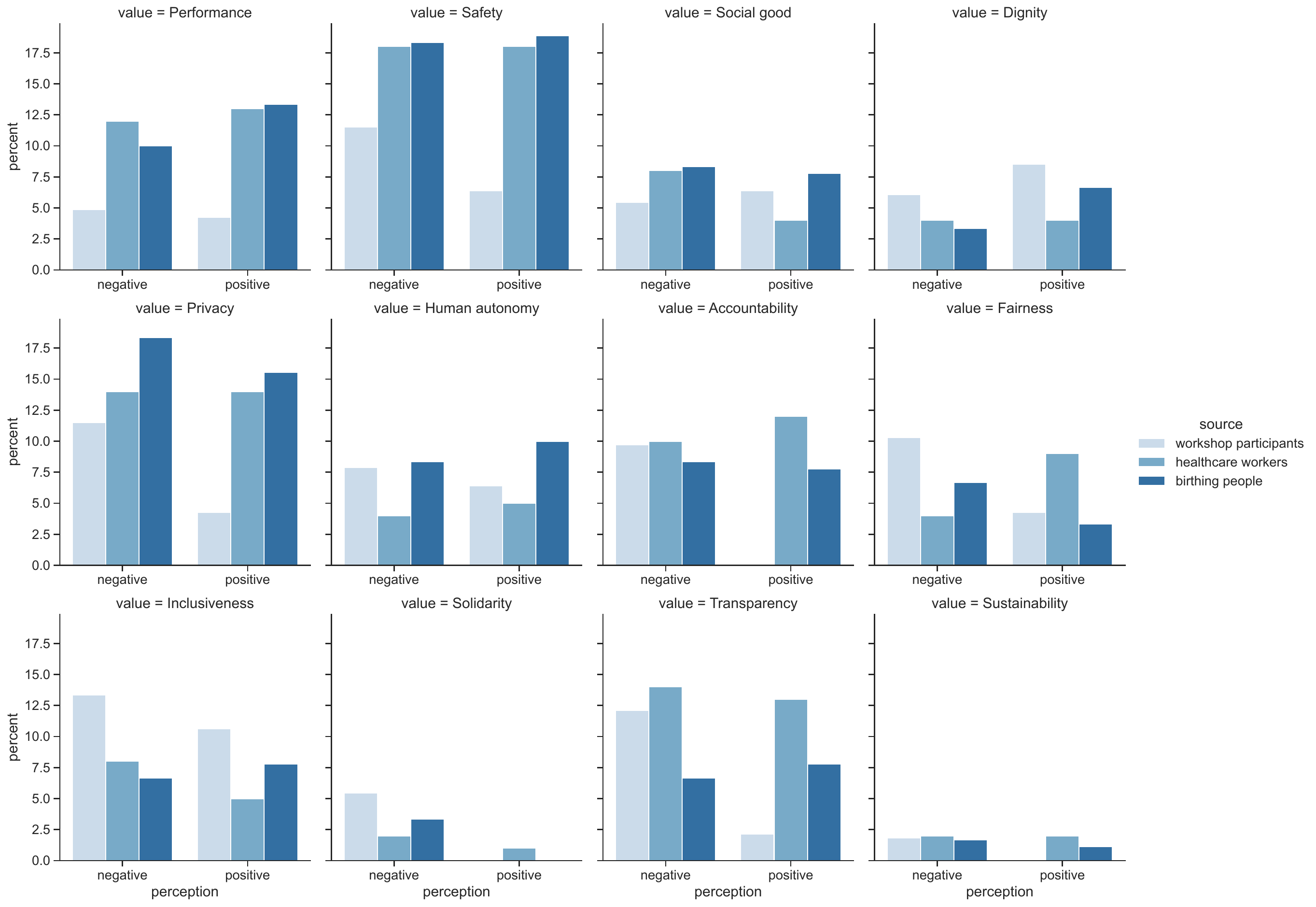}
    \caption{Selection of values broken apart by cohort and general perceptions of AI. \textit{Positive} indicates scores $>3$.
    \newline
    \newline
    \textit{Alt Text: A set of bar plots showing the selection of values, with responses broken apart by participant cohort and general perceptions of AI.}}
\end{figure}

\section{Chatbot Demonstration}
\label{appendix:chatbot-demonstration}

\begin{figure}[H]
    \centering
    \includegraphics[width=0.48\textwidth]{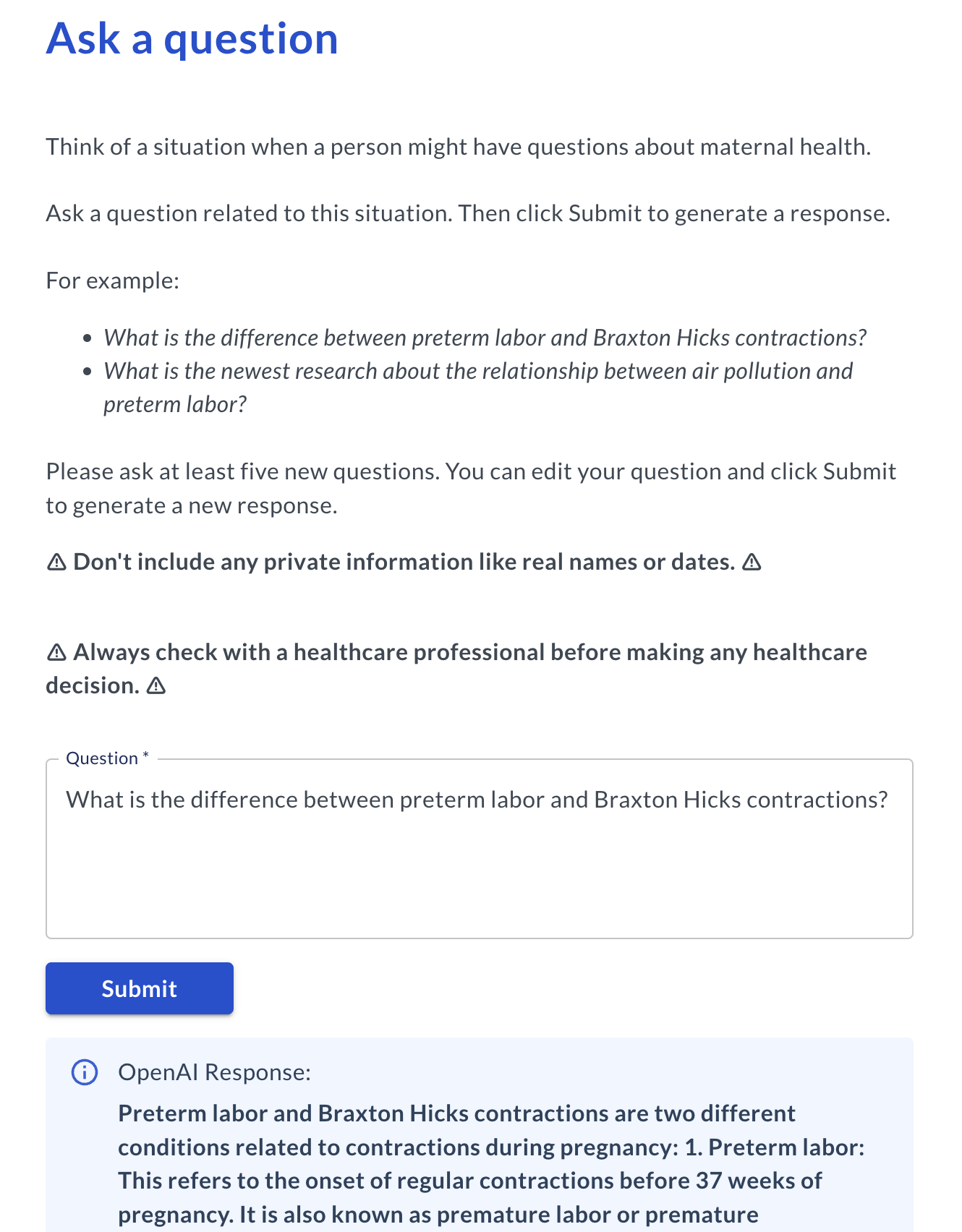}
    \includegraphics[width=0.48\textwidth]{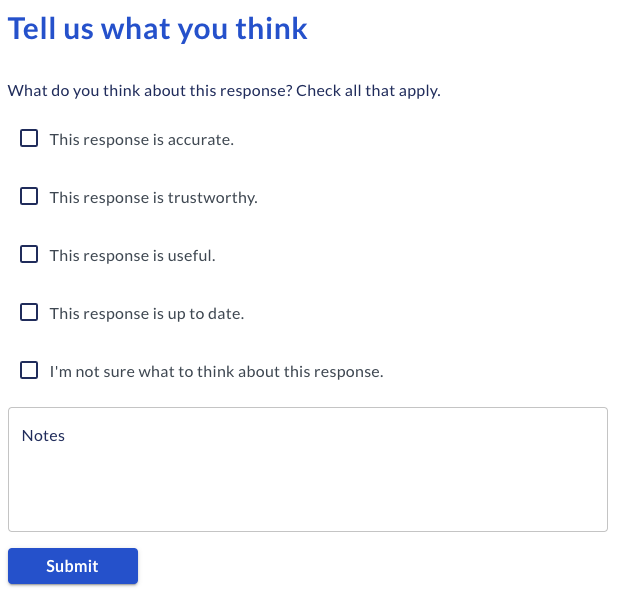}
    \caption{Screenshots of our chatbot demo. Responses were generated using GPT-3.5 from OpenAI. Users could ask multiple questions about maternal health. The workshop demonstration additionally included a free-text entry box for participants to add descriptions of situations in which they might use the chatbot.
    }
    \label{fig:demo}
\end{figure}

\section{Full Survey Questions}
\label{section:full-survey-questions}

\subsection{Consent}

Thank you for participating in this study about the use of AI for maternal healthcare.
This study is being led by Maria Antoniak and Carla S. Alvarado.
The purpose of this research is to gather perceptions about AI tools from a diverse group of people.
This information will be used to create a set of guiding principles for the use of AI for maternal health.
We deeply value your lived experience, and we want to include as many voices as possible in the development of these guiding principles.

\paragraph{Participation}
Your participation in this study is voluntary and anonymous, and you may refuse to participate before the exercise begins, or discontinue at any time with no penalty.

In this study, you will complete a survey describing your experiences with maternal healthcare and your perceptions of AI tools.

\paragraph{Risks \& Benefits}
This study asks about your experiences in maternal health. Any questions related to this experience may make you uncomfortable or bring up feelings of stress. You are always free to decline to answer any question or stop your participation at any time.

You will be paid for your participation in this study according to the rates specific on Prolific. We hope to learn about the various perspectives in maternal health and represent these perspectives in the guidelines we develop, and we hope that these guidelines will shape future AI research for maternal health in ways that are useful, ethical, and appropriate for the community.

\paragraph{Privacy}
We do not collect names or other personally identifiable information. We do collect your Prolific ID, but we do not have any way to link this with your identity. We will not release your Prolific ID to anyone else.

De-identified data from this survey may be shared with the research community at large to advance science and health. We will remove or code any personal information that could identify you before files are shared with other researchers to ensure that, by current scientific standards and known methods, no one will be able to identify you from the information we share. Responses will be stored securely by the Allen Institute for AI with limited access controls to limit exposure to those with a need to know.

\paragraph{Questions}
If you have questions about this study, please reach out to Maria Antoniak or Carla S. Alvarado, or send us a message us on Prolific.

\subsection{Screening Questions}

\begin{enumerate}
    \item What is your Prolific ID?
    \item Do you consent to participate in this study?
    \item \textbf{[Birthing People Only]} Have you given birth in the last five years (2018-2023)? 
    \item \textbf{[Healthcare Workers Only]} Have you ever worked in a healthcare profession?
\end{enumerate}

\subsection{Trust}

\begin{enumerate}
    \item Generally speaking, would you say that most \textbf{people} can be trusted, or that you need to be very careful in dealing with people? \textit{[Likert scale from (1) need to be very careful to (5) most can be trusted]}
    \item Generally speaking, would you say that most \textbf{healthcare providers} can be trusted, or that you need to be very careful in dealing with healthcare providers? \textit{[Likert scale from (1) need to be very careful to (5) most can be trusted]}
\end{enumerate}

\subsection{Familiarity and perceptions of AI}

In the following sections, we'll be asking about your familiarity and perceptions of AI systems. Some examples of AI systems include:
\begin{itemize}
    \item A chatbot used by people to answer general questions
    \item An AI system used by a medical clinic to predict whether a patient has a disease
    \item An AI system used by a bank to predict whether an applicant will repay a loan
    \item An AI system used by a marketing company to match ads to viewers
    \item An AI system used by a streaming company to recommend movies to users
\end{itemize}

\begin{enumerate}
    \item How familiar are you with NLP, machine learning, and/or AI? \textit{[Not at all familiar, Heard of it but never sought more information, Read about it, Occasionally used tools relying on these methods, Extensively used tools relying on these methods, Trained or finetuned my own models]}
    \item Have you ever used an AI chatbot like ChatGPT? \textit{[Likert scale from (1) never to (5) all the time]}
    \item AI can eliminate a lot of tedious work for people. \textit{[Likert scale from (1) strongly disagree to (5) strongly agree]}
    \item The overuse of AI may be harmful and damaging to humans. \textit{[Likert scale from (1) strongly disagree to (5) strongly agree]}
    \item Life will be easier and faster with AI. \textit{[Likert scale from (1) strongly disagree to (5) strongly agree]}
    \item AI turns people into just another number. \textit{[Likert scale from (1) strongly disagree to (5) strongly agree]}
\end{enumerate}

\subsection{Values}

\textit{These definitions are drawn directly from \citet{jakesch2022different}.}

\paragraph{Fairness} A fair NLP system treats all people equally. Developers of fair NLP systems ensure, as far as possible, that the system does not reinforce biases or stereotypes. A fair system works equally well for everyone independent of their race, gender, sexual orientation, and ability.

\paragraph{Privacy} An NLP system that respects people’s privacy implements strong privacy safeguards. Developers of privacy-preserving NLP systems minimize, as far as possible, the collection of sensitive data and ensure that the NLP system provides notice and asks for consent.

\paragraph{Sustainability} A sustainable NLP system preserves the environmental quality of current and future generations. Developers of sustainable NLP systems minimize, as far as possible, electricity use and reduce waste.

\paragraph{Inclusiveness} Inclusive NLP systems empower everyone and engage all people. Developers of inclusive AI systems  consider, as far as possible, the needs of people who might otherwise be excluded or marginalized.

\paragraph{Safety} A safe NLP system performs reliably and safely. Developers of safe NLP systems implement strong safety measures. They anticipate and mitigate, as far as possible, physical, emotional, and psychological harms that the system might cause.

\paragraph{Social good} An NLP system that promotes social good supports, as far as possible, human well-being and flourishing, peace and happiness, and the creation of socio-economic opportunities.

\paragraph{Dignity} An NLP system that respects human dignity upholds the inherent worth of every individual.  In addition to respective legislation, developers ensure that the system respects human rights and does not diminish human dignity.

\paragraph{Performance} A high-performing NLP system consistently produces good predictions, inferences or answers. Developers of high-performing NLP systems ensure, as far as possible, that the system’s results are useful, accurate and produced with minimal delay.

\paragraph{Accountability} An accountable NLP system has clear attributions of responsibilities and liability. Developers and operators of accountable AI systems are, as far as possible, held responsible for their impacts. An accountable system also implements mechanisms for appeal and recourse.

\paragraph{Transparency} A transparent NLP system produces decisions that people can understand. Developers of transparent AI systems ensure, as far as possible, that users can get insight into why and how a system made a decision or inference.

\paragraph{Human autonomy} An NLP system that respects people’s autonomy avoids reducing their agency. Developers of autonomy-preserving AI systems ensure, as far as possible, that the system provides choices to people and preserves or increases their control over their lives.

\paragraph{Solidarity} A solidary NLP system does not increase inequality and leaves no one behind. Developers of solidary AI systems ensure, as far as possible, that the prosperity as well as the burdens created by NLP are shared by all.

\begin{enumerate}
    \item Please select any five values from the list below that you think are most important for NLP systems for maternal health. \textit{[Fairness, Privacy, Sustainability, Inclusiveness, Safety, Social good, Dignity, Performance, Accountability, Transparency, Human autonomy, Solidarity]}
    \item (Optional) Was anything missing from the list of values above? If so, describe here.
\end{enumerate}

\subsection{Information Seeking Behavior}

\begin{enumerate}
    \item When I have questions about maternal health, I often turn to this resource to assist me. \newline \textit{[Likert scale from (1) strongly disagree to (5) strongly agree for each of the following options]}
    \begin{itemize}
        \item Peer-reviewed scientific publications (for example: Journal of the American Medical Association)
        \item Online communities and forums (for example: Facebook groups, Reddit groups)
        \item Social media (for example: Instagram, TikTok, YouTube)
        \item Government resources (for example: FDA, CDC websites and announcements)
        \item News articles (for example: CNN, WSJ)
        \item Textbooks, books
        \item Expert-written websites (for example: WebMD, Mayo Clinic)
        \item Curated medical resources (for example: UpToDate)
    \end{itemize}
    \item (Optional) Outside of the resources listed above, are there are other resources you turn to for help in answering questions about maternal healthcare?
    \item When I have questions about maternal health, I often turn to this person to assist me.  \newline \textit{[Likert scale from (1) strongly disagree to (5) strongly agree for each of the following options]}
    \begin{itemize}
        \item OB/GYN
        \item Nurse
        \item Midwife
        \item Doula
        \item Community health worker
        \item Healthcare reseracher
        \item Government workers
        \item Friends \& family
        \item  Colleagues
    \end{itemize}
    \item (Optional) Outside of the people listed above, are there are other people you turn to for help in answering questions about maternal healthcare?
\end{enumerate}

\subsection{Effects of NLP/AI}

\begin{enumerate}
    \item This person would benefit from an NLP/AI chatbot in their work. \newline \textit{[Likert scale from (1) strongly disagree to (5) strongly agree for each of the following options]}
    \begin{itemize}
        \item OB/GYN
        \item Nurse
        \item Midwife
        \item Doula
        \item Community health worker
        \item Healthcare reseracher
        \item Government workers
        \item Friends \& family
        \item  Colleagues
    \end{itemize}
    \item \textbf{[Workshop Only]} What impact do you expect an NLP chatbot would have on your work? \textit{[Likert scale from (1) mostly harms to (5) mostly benefits]}
    \item \textbf{[Birthing People Only]}  I would want my maternal healthcare provider to use AI systems. \textit{[Likert scale from (1) strongly disagree to (5) strongly agree]}
    \item \textbf{[Birthing People Only]} I would want my maternal healthcare provider to tell me if they use AI systems. \textit{[Likert scale from (1) strongly disagree to (5) strongly agree]}
    \item \textbf{[Birthing People Only]} I would want my maternal healthcare provider to tell me if they use AI systems. \textit{[Likert scale from (1) strongly disagree to (5) strongly agree]}
    \item \textbf{[Birthing People Only]} How often do/did you consult internet resources for pregnancy-related questions during your pregnancy? \textit{[Likert scale from (1) never to (5) frequently]}
    \item \textbf{[Birthing People Only]} If so, what were these pregnancy-related questions about? (check all that apply) \textit{[medical worries/concerns during pregnancy, mental health worries/concerns during pregnancy, the pregnancy process in general, labor and delivery, birth control option after delivery, the postpartum process, taking care of a newborn, medical worries/concerns after pregnancy, mental health worries/concerns after pregnancy, Other (write in)]}
\end{enumerate}

\subsection{Demographics}

\textit{All of these questions were optional.}

\begin{enumerate}
    \item Which gender(s) do you identify with? \textit{[woman, man, non-binary, prefer not to disclose, other (write in)]}
    \item What is your age range? \textit{[$<$18, 18-24, 25-34, 35-44, 45-54, 55-64, 65-74, 74+]}
    \item Describe your race/ethnicity (check all that apply) \textit{[African-American/Black, Middle Eastern/North African, Native American/Alaska Native/First Nations, Pacific Islander, Hispanic/Latino/a/x, East Asian (including Chinese, Japanese, Korean, Mongolian, Tibetan, and Taiwanese), South Asian (including Bangladeshi, Bhutanese, Indian, Nepali, Pakistani, and Sri Lankan), Southeast Asian (including Burmese, Cambodian, Filipino, Hmong, Indonesian, Laotian, Malaysian, Mien, Singaporean, Thai, and Vietnamese), White, Other (write in)]}
    \item Where do you live in the U.S.? \textit{[Northeast, Midwest, South, West, I do not live in the U.S.]}
    \item \textbf{[Workshop Only]} What is your education level? \textit{[Trade school (for example: AA, AS), College (for example: BA, BS), Master's degree (for example: MS, MA, MPH), Medical degree (for example: MD, DO), Other professional degree (for example: JD), PhD]}
    \item \textbf{[Workshop Only]} Describe your professional background. Check every profession that has \textbf{ever} applied to you. \textit{[Community health worker/Promotoras, Community-based organization (non-profit), Doula, Certified Midwife, Certified Nurse Midwife, L\&D Nurse, OB/GYN, Other perinatal health care provider, Other perinatal support (non-health care) provider, Non-perinatal health care provider, Health care management/administration, Health care services researcher, Population Health/Public Health researcher, Local/State public health entity, Federal Government employee in public health (for example: CDC, HHS, CMS and other related units)], AI / machine learning / NLP researcher or engineer, Other (write in)}
    \item \textbf{[Workshop Only]} Have you ever sought maternal or reproductive support from a healthcare provider (for example: pregnancy support, contraception support)? \textit{[yes, no, unsure, not applicable]}
    \item \textbf{[Prolific Only]} Do you or have you ever worked in healthcare (in a hospital or clinic, as a community health worker, as a public health researcher, or in any other capacity)? \textit{[yes, no]}
    \item  \textbf{[Prolific Only]}  Do you or have you ever worked in maternal/perinatal healthcare (as a nurse, midwife, researcher, or in any other capacity)? \textit{[yes, no]}
    \item Feel free to share any additional comments or feedback about this survey here. If anything was confusing or unclear, we'd love to know so that we can improve this survey in the future.
\end{enumerate}

\end{document}